\documentclass[american,english,wcp]{jmlr}
\usepackage[T1]{fontenc}
\usepackage[latin9]{inputenc}
\usepackage{amsbsy}
\usepackage{amstext}
\usepackage{amssymb}
\usepackage{graphicx}
\PassOptionsToPackage{normalem}{ulem}
\usepackage{ulem}

\makeatletter

\providecommand{\tabularnewline}{\\}

\usepackage{longtable}
\usepackage{booktabs}
\usepackage{lineno}

\usepackage{colortbl} 
\definecolor{header_color}{rgb}{0.74,0.88,0.91}
\definecolor{even_color}{rgb}{0.9,0.9,0.9}
\definecolor{subheader_color}{rgb}{0.85,0.93,0.95}
\definecolor{childheader_color}{rgb}{1.0,0.93,0.87}

\makeatletter
\let\Ginclude@graphics\@org@Ginclude@graphics 
\makeatother

\jmlryear{2026}
\jmlrworkshop{ACML 2026}
\jmlrproceedings{PMLR}{Proceedings of Machine Learning Research}
\editors{Andy Song, Bo Han and Sarah Erfani}

\author{%
\Name{Tu Anh Hoang Nguyen}\thanks{Equal contributions as co-first authors.}
\Email{joseph.nguyen@deakin.edu.au}\\
\Name{Dang Nguyen}\footnotemark[1]
\Email{d.nguyen@deakin.edu.au}\\
\Name{Tri-Nhan Vo}
\Email{s223032975@deakin.edu.au}\\
\addr Applied Artificial Intelligence Initiative (A$^2$I$^2$), Deakin University, Australia
\AND
\Name{Thuc Duy Le}
\Email{Thuc.Le@Adelaide.edu.au}\\
\addr School of Computer Science and Information Technology, Adelaide University, Australia
\AND
\Name{Trung Le}
\Email{trunglm@monash.edu}\\
\addr Department of Data Science \& AI, Monash University, Australia
\AND
\Name{Sunil Gupta}
\Email{sunil.gupta@deakin.edu.au}\\
\addr Applied Artificial Intelligence Initiative (A$^2$I$^2$), Deakin University, Australia
}

\title[Causal-Aware Tabular GANs]{Causal-Aware Tabular GANs with Reinforcement Learning}

\makeatother

\usepackage{babel}
\begin{document}
\makeatletter
\let \@jmlrpages \@empty
\makeatother
\maketitle
\begin{abstract}
Existing tabular data generation methods primarily focus on matching
statistical distributions between real and synthetic data, often overlooking
the preservation of underlying causal relationships. As a result,
generated samples may appear realistic while failing to maintain the
causal structure required for reliable downstream analysis. We propose
\textbf{CA-GAN}, a causal-aware generative framework for tabular data
synthesis that explicitly incorporates causal knowledge into both
the training and generation processes. CA-GAN first extracts a causal
graph from real data to provide structural prior knowledge, then employs
a graph-conditioned Conditional WGAN-GP whose sub-generators model
variables according to their causal dependencies. More importantly,
we introduce a reinforcement learning-based objective that treats
causal graph discrepancy between real and synthetic data as a reward
signal, enabling causal consistency to become an explicit optimization
target during training rather than an implicit consequence of sampling
order. Extensive experiments on 14 synthetic and real-world datasets
demonstrate that CA-GAN consistently outperforms seven state-of-the-art
baselines in causal preservation while achieving strong downstream
utility, privacy preservation, and data quality. These results show
that CA-GAN provides an effective and practical solution for generating
high-quality synthetic tabular data that better respects underlying
causal mechanisms.
\end{abstract}

\section{Introduction}

Tabular data is one of the most common data types in various fields
such as healthcare and finance. However, many machine learning (ML)
applications are restricted due to the limitations of real-world datasets
such as quality, privacy, and utility. Therefore, the generation of
synthetic tabular data has emerged as a promising direction \cite{challagundla2025synthetic}.

Several generative approaches have been proposed for synthesizing
tabular data, including models based on Generative Adversarial Networks
(GANs) \cite{Xu2019}, Variational Autoencoders (VAEs) \cite{Xu2019},
Diffusion \cite{Shi2025}, and recently, Large Language Models (LLMs)
\cite{Borisov2023,zhang2023generative,nguyen2024tabular}. Early methods
such as CTGAN \cite{Xu2019} use GANs as they effectively capture
complex data distributions. Others use VAEs, exemplified by TVAE \cite{Xu2019},
offer a probabilistic framework that enables efficient latent representation
learning and robust reconstruction, making them valuable for various
tabular domains. Recently, LLM-based methods like Great \cite{Borisov2023},
TapTap \cite{zhang2023generative}, and Pred-LLM \cite{nguyen2024tabular}
have emerged as promising candidates for tabular generation methods,
thanks to their extensive capabilities such as arbitrary conditioning
and correlation capturing. However, LLM-based methods come with significant
limitations such as great computation cost and slow inference. Recent
studies \cite{yang2025doubling} show that CTGAN provide a faster
and more resource-efficient alternative, with training in roughly
a minute, compared with hours for Great, making GAN-based methods
as practical choice for real-world applications that demand rapid
and scalable data generation.

While current generative methods for tabular data primarily focus
on matching statistical distributions between original and synthetic
datasets, they often overlook the preservation of underlying \textit{causal
relationships}. As a result, synthetic data may appear realistic but
yield misleading inferences under intervention. In biomedicine, breaking
causal links between therapies, biomarkers, and outcomes can distort
conclusions about treatment efficacy--posing risks in fields like
cancer and blood pressure studies \cite{goncalves2020generation,shi2022generating}.
In economics, overlooking causal pathways may lead to incorrect policy
decisions such as misjudging the effect of education reforms on income
\cite{yang2024synthetic}. These limitations necessitate novel frameworks
in tabular data generation, which are \textbf{causality-preserving}.

Although causal preservation is increasingly recognized as essential
in synthetic tabular data generation \cite{Bauwelinckx2025,Jiang2026},
only a few methods have explicitly incorporated this aspect in the
data generation process. As far as we know, only two methods CausalGAN
\cite{kocaoglu2017causalgan} and Causal-TGAN \cite{wen2022causal}
use a \textit{directed acyclic graph} (DAG) to guide the data generation.
While CausalGAN applies this idea to binary image attributes, Causal-TGAN
is the first method to target tabular data. In Causal-TGAN, a predefined
DAG is used to determine the feature/variable sampling order, ensuring
that the generated samples follow a plausible causal structure. However,
its training process remains identical to conventional GANs, which
relies only on distinguishing real from fake samples without encouraging
causal alignment. Consequently, its generator may not generate synthetic
data that fully capture the causal relation as in the original data.

To address this limitation, we propose \textbf{CA-GAN} (\textit{Causal-aware
GAN}), a robust model that integrates causal awareness in both training
and generation phases. First, we use a causal discovery algorithm
to extract the \textit{base causal graph} from the original data and
consider it as prior causal information. Following Causal-TGAN, we
design our GAN architecture with multiple sub-generators and one discriminator.
Each sub-generator corresponding to one node in causal graph models
a conditional distribution, which receives not only a latent noise
as input but also the values of its parent nodes. Finally, to train
our generator, besides the standard adversarial loss term, we introduce
a novel loss term that minimizes the discrepancy between the base
causal graph and the causal graph extracted from fake data. This discrepancy
is measured by Structural Hamming Distance (SHD), which is a \textit{non-differentiable}
metric. We circumvent this challenge with a reinforcement learning
(RL) approach, specifically a policy gradient (REINFORCE) method that
treats SHD score as a \textit{reward} signal -- enabling our generator
to update its parameters based on the returned rewards. 

We summarize \textbf{our contributions} as follows:

(1) \textbf{Proposed framework:} We propose CA-GAN model for tabular
data generation, which aims to preserve causal relationships in the
original data.

(2) \textbf{Algorithmic innovations:} We introduce a RL-based training
objective for our generator, which aligns the causal structures of
original and generated data.

(3) \textbf{Empirical validation:} We conduct extensive experiments on
14 synthetic and real-world datasets to show CA-GAN delivers superior
causal preservation, downstream utility, privacy protection, and data
quality.

(4) \textbf{Novelty:} While Causal-TGAN uses a DAG to guide feature
sampling order, its training process remains a conventional GAN focusing
purely on real vs. fake discrimination. \textit{It lacks theoretical
guarantees for causal alignment and does not penalize the model for
breaking causal links}. In contrast, CA-GAN incorporates causal structure
directly into the \textbf{training objective} by optimizing an SHD-based
signal via RL. This introduces a fundamental shift: causal consistency
is no longer an implicit inductive bias but an \textbf{explicit optimization
target} during training.

\section{Related Works}

\textbf{Generative models for tabular data:} Synthetic tabular data
is increasingly used to add diversity, support validation, reduce
privacy risk, and enable large-scale experiments \cite{challagundla2025synthetic}.
Recently, generative models have been adapted to tabular data. The
early-stage dominant line is GAN and VAE-based methods, including
CTGAN and TVAE \cite{Xu2019}. Beyond that, TabDiff \cite{Shi2025}
is built on Diffusion, while recent methods such as Great \cite{Borisov2023},
TapTap \cite{zhang2023generative}, and Pred-LLM \cite{nguyen2024tabular}
leverage LLMs. LLM-based methods provide two core gains: arbitrary
conditioning on any feature subset without re-training and context-aware
semantics that capture feature correlation. However, these benefits
come with notable trade-offs that limit their practical deployment
-- especially in computation and numerical fidelity. First, their
billions-of-parameters size is extremely resource-demanding, require
huge amount of time and computation resources for both training and
inference. Second, tokenization in LLMs are primarily designed for
natural language, not for structured or numerical data. These constraints
make LLM-based models less suitable for resource-limited settings
or applications demanding fast inference \cite{yang2025doubling}.

\textbf{Causality in tabular data:} Causal structure is crucial in
synthesizing tabular data since breaking cause-effect links can skew
downstream conclusions \cite{Bauwelinckx2025,Jiang2026}. For example,
in biomedicine, it can misstate treatment effects \cite{goncalves2020generation}
and in economics, it can mislead policy analysis \cite{yang2024synthetic}.
Structural Causal Models (SCMs) and the potential-outcomes approach
provide clear, rigorous tools to represent and test these relationships
\cite{Pearl2009}. Building on these ideas, few generative models
combine causality with synthesis. CausalGAN \cite{kocaoglu2017causalgan}
uses a causal graph to steer image attribute generation and permit
interventions on them. DECAF \cite{VanBreugel2021} aligns sampling
with a causal view of sensitive attributes to promote fairness. Causal-TGAN
\cite{wen2022causal} generates tabular variables in causal graph
order while keeping the standard adversarial loop. Namely, Causal-TGAN
conditions each variable on its parents in the causal graph and organizes
sub-generators accordingly, extending CTGAN \cite{Xu2019} without
altering the core discriminator-generator objective functions. So
far, only Causal-TGAN is a causal-aware generative model applied to
tabular data. While its graph-guided ordering promotes causal consistency
during sampling, its learning objective remains that of a conventional
GAN focused on real-fake discrimination rather than causal alignment,
which may fail to preserve causal semantics.

\section{Framework}

\subsection{Problem Definition}

Given a real tabular dataset $\mathcal{D}_{real}=\{x_{i}\}_{i=1}^{N}$,
each row is a sample $x_{i}$ with $M$ features $\{X_{1},\dots,X_{M}\}$.
Its underlying causal structure represented by a \textit{directed
acyclic graph} (DAG) is $\mathcal{G}_{real}=(V,E)$, where $V=\{X_{1},...,X_{M}\}$
is a set of nodes corresponding to $M$ features in the dataset ${\cal D}_{real}$
and $E$ is a set of edges such that a node $X_{i}$ is a parent of
node $X_{j}$ if there is a directed edge from $X_{i}$ to $X_{j}$. 

Our goal is to learn a generator $G_{\theta}$ capable of producing
a synthetic dataset $\mathcal{D}_{fake}=\{\hat{x}_{i}\}_{i=1}^{N}$
that is both \textbf{statistically realistic} and \textbf{causally
faithful}.

\subsection{Proposed method CA-GAN}

\begin{figure}[!t]
\centering
\begin{centering}
\includegraphics[width=0.8\columnwidth]{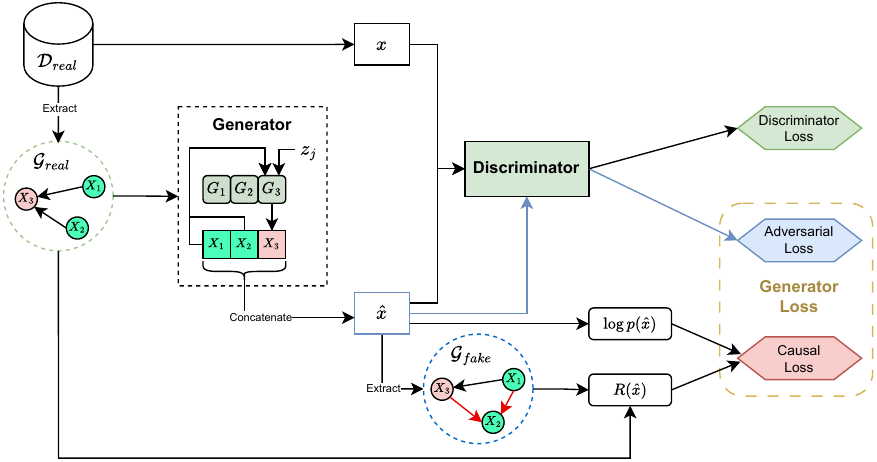}
\par\end{centering}
\caption{\label{fig:Our-framework-CA-GAN.}Our framework CA-GAN. First, we
extract the causal graph ${\cal G}_{real}$ from ${\cal D}_{real}$.
Next, we train a WGAN-GP model with $\{G_{1},...,G_{M}\}$ sub-generators
and one discriminator $D$. Each $G_{j}$ synthesizes variable $X_{j}$
sequentially following the node ordering in $\mathcal{G}_{real}$.
The input of $G_{j}$ includes parent values $\mathbf{PA}_{j}$ and
noise $z_{j}$. We train $D$ with a standard \textit{discriminator
loss} using fake samples $\hat{x}$ and real samples $x$. Finally,
we design a reward $R(\hat{x})$ via the Structural Hamming Distance
between ${\cal G}_{real}$ and ${\cal G}_{fake}$ extracted from $\hat{x}$.
Using the Policy Gradient theorem, we construct the \textit{causal
loss} -- the product of the log-probability of the generator output
$\log p(\hat{x})$ and $R(\hat{x})$. We train the generator with
both\textit{ adversarial loss }and\textit{ causal loss}.}
\end{figure}

As conventional GAN-based synthesizers focus on matching the data
distributions, the causal dependencies among features/variables are
often overlooked, leading to generated samples with inconsistent causal
relationships. To address this problem, CA-GAN integrates a REINFORCE-based
causal regularization that optimizes the generator through both adversarial
loss and causal rewards, minimizing the structural discrepancy between
the graph inferred from $\mathcal{D}_{fake}$ and that inferred from
$\mathcal{D}_{real}$, while preserving the overall distributional
fidelity $p({\cal D}_{fake})\approx p({\cal D}_{real})$. This process
is illustrated in Figure \ref{fig:Our-framework-CA-GAN.} and detailed
below.

\subsubsection{Causal knowledge extraction}

To incorporate causal dependencies into the training and generation
phases, CA-GAN begins by estimating the underlying causal graph $\mathcal{G}_{real}$ from
original real dataset ${\cal D}_{real}$. This graph captures how
variables influence one another, and forms the structural backbone
that guides the generator to produce synthetic data consistent with
true cause-effect relationships rather than simple correlations. We
employ the \textit{PC}\textbf{\textit{ }}\textit{algorithm} \cite{spirtes2000causation}
to infer $\mathcal{G}_{real}$ through conditional independence testing.
PC iteratively removes and orients edges to yield a completed partially
directed acyclic graph (CPDAG). From the resulting CPDAG, a corresponding
DAG or \textit{causal graph} $\mathcal{G}_{real}$  is obtained. Given
a node (variable) $X_{j}$, $\mathcal{G}_{real}$ provides its parent
set $\mathbf{PA}_{j}$ and a valid topological ordering of nodes.
These serve as the structural information for CA-GAN, ensuring that
each variable $X_{j}$ is generated conditionally on its causal parents.

Moreover, the data in $\mathcal{D}_{real}$ are generated from a \textit{Structural
Causal Model} (SCM) $\mathcal{M}=\langle\mathcal{X},\mathcal{E},\mathcal{F}\rangle$,
where ${\cal X}=\{X_{1},...,X_{M}\}$ denotes the features, $\mathcal{E}=\{E_{1},\dots,E_{M}\}$
represents exogenous noises, and $\mathcal{F}=\{f_{1},\dots,f_{M}\}$
is a set of deterministic structural functions. Each feature is generated
as $X_{j}=f_{j}(\mathbf{PA}_{j},E_{j})$, where $E_{j}\sim{\cal N}(0,1)$.
Although PC assists us in finding the causal relationships among the
data features organized in the causal graph $\mathcal{G}_{real}$,
it does not provide the set of deterministic structural functions
$\mathcal{F}$ which is crucial to generate causal-following synthetic
samples. Our aim is to develop a causal-aware generator $G=\{G_{j}\}_{j=1}^{M}$
simulated the underlying SCM where each sub-generator $G_{j}$ aims
to approximate the corresponding structural function $f_{j}$.

\subsubsection{Generator architecture and motivated theory}

It is crucial to design a causal-aware generator $G$ exploiting the
causal graph $\mathcal{G}_{real}$. Our generator consists of multiple
sub-generators $G=\{G_{j}\}_{j=1}^{M}$, where each $G_{j}$ models
the conditional distribution of variable $X_{j}$ given its parent
set $\mathbf{PA}_{j}$ and an independent noise vector $z_{j}\sim\mathcal{N}(0,1)$.
Sampling proceeds auto-regressively along the topological order of
$\mathcal{\mathcal{G}}_{real}$, ensuring that parent values are available
before generating each node.

\selectlanguage{american}%
Following \foreignlanguage{english}{\cite{dang2026@causal}, }we leverage
a theoretical result showing that under causal factorization and additive
transport costs, minimizing the \foreignlanguage{english}{\textit{Wasserstein
distance} (WS distance)} encourages alignment between generated mechanisms
and structural functions. Without loss of generality, we assume the
order $1,...,M$ and the cost function $c$ satisfying the \textit{additive
property}: $c([X_{A},X_{B}],[X'_{A},X'_{B}])=c(X_{A},X'_{A})+c(X_{B},X'_{B})$
where $A,B\subset\{1,...,M\}$ are disjoint with $X_{C}=[X_{i}]_{i\in C}$.
Obviously, $c\left(X_{C},X'_{C}\right)=\sum_{i\in C}\left|X_{i}-X_{i}^{'}\right|^{p}$
with $p>0$ satisfies the additive property.

\selectlanguage{english}%
\begin{theorem}
\label{thm:main} Under an additive cost function $c$, the WS distance
$\mathcal{W}_{c}\left(p_{{\cal D}},p_{G}\right)$ has a lower-bound:
$\mathcal{W}_{c}\left(p_{{\cal D}},p_{G}\right)\geq\sum_{j=1}^{M}d_{j}\left(G_{j},f_{j}\right)$.
The distances $d_{j}$ are defined as: $d_{j}\left(G_{j},f_{j}\right)=\min_{\gamma_{j}\in\Gamma_{j}}\mathbb{E}_{\gamma_{j}}\left[c\left(f_{j}\left(X_{\mathbf{PA}_{j}},E_{j}\right),G_{j}\left(X_{\mathbf{PA}_{j}^{G}},E_{j}^{G}\right)\right)\right]$,
where $\Gamma_{j}=\Gamma\left(\alpha_{j},\alpha_{j}^{G}\right)$ is
the set of couplings over $\alpha_{j}$ and $\alpha_{j}^{G}$, $\alpha_{j}$
consists of samples formed by exogenous noises $E_{j}$ and $X_{\mathbf{PA}_{j}}$
sampled from the SCM of $\mathcal{G}_{real}$, and $\alpha_{j}^{G}$
is defined the same but with respect to the SCM induced by the causal-aware
generator $G$.
\end{theorem}

\selectlanguage{american}%
While Theorem \foreignlanguage{english}{\ref{thm:main}} establishes
that our WGAN-GP adversarial objective forces local mechanism alignment,
it operates under the strict assumption that the generator's topological
ordering perfectly matches the true causal structure. If the structural
dependencies are flawed, minimizing the Wasserstein distance will
inadvertently force $G_{j}$ to learn spurious, non-causal correlations.
This theoretical limitation necessitates our novel dual-objective
design: we must simultaneously introduce a structural penalty (our
RL-based SHD reward) to ensure the variables interact through the
correct graph skeleton, allowing the adversarial Wasserstein minimization
to operate on a valid causal topology.
\selectlanguage{english}%

\subsubsection{Training our approach with WGAN-GP}

As suggested by our theory, we need to minimize the WS distance $\mathcal{W}_{c}\left(p_{{\cal D}},p_{G}\right)$.
To this end, we adopt the WGAN model \cite{Arjovsky2017} for our
generator but factorize it in a way that is tailored to the causal
graph $\mathcal{G}_{real}$ and simultaneously train a discriminator
$D$.

\paragraph{Training discriminator $D$}

CA-GAN trains its generator-discriminator pair under the WGAN-GP \cite{Gulrajani2017}
objective to ensure stable adversarial learning and accurate distribution
matching between real and fake data. The discriminator $D$ is trained
to assign higher scores to real samples and lower scores to generated
ones, effectively estimating the WS distance between the real and
fake distributions. The \textit{discriminator loss} is:
\begin{equation}
{\displaystyle \mathcal{L}_{D}=-\mathbb{E}_{x\sim p_{{\cal D}}}[D(x)]+\mathbb{E}_{\hat{x}\sim p_{G}}[D(\hat{x})]+GP,}\label{eq:discriminator-loss}
\end{equation}
where $D(x)$ and $D(\hat{x})$ denote critic scores for real and
fake samples. The gradient penalty term ($GP$) from WGAN-GP is included
to enforce smoothness and stability during training, ensuring the
discriminator remains well-behaved while it distinguishes real samples
from fake samples. It is defined as: $GP=\mathbb{E}_{\tilde{x}\sim p_{\tilde{x}}}[(||\nabla_{\tilde{x}}D(\tilde{x})||_{2}-1)^{2}]$,
where $\tilde{x}$ are random points along straight lines between
real and fake samples i.e., $\tilde{x}=\epsilon x+(1-\epsilon)\hat{x}$,
$\epsilon\sim{\cal U}(0,1)$.

\paragraph{Training generator $G$}

During generator updates, CA-GAN synthesizes variables sequentially
based on the node ordering in the causal graph $\mathcal{G}_{real}$,
producing a batch of fake samples $\hat{x}$. Note that each sample
$\hat{x}_{i}$ is a concatenation of all generated variables. These
fake samples are used solely for adversarial updates of the discriminator.
The generator is trained to increase the critic\textquoteright s scores
on these generated samples, effectively narrowing the gap between
the model distribution $p_{G}$ and the real data distribution $p_{{\cal D}}$.
Its \textit{adversarial loss} follows the standard WGAN formulation:
\begin{equation}
\mathcal{L}_{adv}=-\mathbb{E}_{\hat{x}\sim p_{G}}[D(\hat{x})]\label{eq:adversarial-loss}
\end{equation}
By minimizing this loss, the generator improves the realism of the
produced samples under the critic\textquoteright s measure.

\paragraph{Causality-preserving objective for generator $G$}

While the adversarial training enables CA-GAN to achieve high distributional
realism, it alone does not ensure that generated data respect the
underlying causal relationships among variables. To overcome this
limitation, we introduce a causality-preserving loss and integrate
it into the final \textit{generator loss}:
\begin{equation}
{\cal L}_{G}={\cal L}_{adv}+\lambda\cdot{\cal L}_{causal}\label{eq:generator-loss}
\end{equation}
Here, ${\cal L}_{adv}$ enforces distributional realism through adversarial
training while ${\cal L}_{causal}$ introduces reinforcement-based
causal feedback to encourage structural consistency with the real
causal graph. The coefficient $\lambda$ controls the balance between
these two loss terms--ensuring that causal regularization strengthens
structural fidelity without destabilizing adversarial training. This
combined formulation enables our generator to learn both \textit{statistical
fidelity} and \textit{causal correctness} within a single optimization
framework.

To encourage the generator to preserve causal relationships, we define
the causal loss term ${\cal L}_{causal}$ as a \textit{structural
reward} based on the difference between causal graphs derived from
real and fake samples. Specifically, the Structural Hamming Distance
(SHD) metric is used to measure the difference between the adjacency
matrices of the real causal graph $\mathcal{G}_{real}$ and the fake
causal graph $\mathcal{G}_{fake}$ (inferred from fake samples $\hat{x}$
using the PC algorithm). The reward for fake samples $\hat{x}$ is:
\begin{equation}
R(\hat{x})=-\mathrm{SHD}\big(\mathcal{G}_{real},\mathcal{G}_{fake}\big),\label{eq:structural-reward}
\end{equation}
where ${\cal G}_{fake}=\text{PC}(\hat{x})$. A higher reward (i.e.,
less negative) indicates closer structural alignment. We then define
the \textit{causal loss} term as:
\begin{equation}
{\cal L}_{causal}=-\mathbb{E}_{\hat{x}\sim p_{G}}[R(\hat{x})]\label{eq:causal-loss}
\end{equation}
However, this loss is \textit{non-differentiable} with respect to
the generator parameters since SHD operates on discrete graph structures.
To optimize this loss effectively, our method CA-GAN builds upon the
core theoretical result of Williams's 1992 paper \cite{Williams1992},
which demonstrates that the expected weight update in a REINFORCE
network is proportional to the gradient of the expected reinforcement
i.e., $\mathbb{E}[\bigtriangleup W]\propto\nabla_{W}\mathbb{E}\{r\mid W\}$.
We adapt this principle (aka \textit{Policy Gradient theorem}) to
our problem. Here, the generator $G$ is treated as a stochastic policy.
The REINFORCE network parameters $W$ are replaced by the generator
parameters $\theta$, which govern how the generator learns to generate
samples $\hat{x}$ (i.e., the action). Likewise, the reward signal
$r$ is interpreted as the \textit{structural} reward $R(\hat{x})$.
Under this reformulation, the theorem naturally yields the gradient
of the causal loss:
\begin{equation}
\nabla_{\theta}{\cal L}_{causal}=-\mathbb{E}_{\hat{x}\sim p_{G}}\Big[R(\hat{x})\nabla_{\theta}\log p_{\theta}(\hat{x})\Big]\label{eq:causal-loss-derivative}
\end{equation}
which expresses that the generator $G$ updates its parameters $\theta$
in the manner of increasing the likelihood of the outputs (i.e., fake
samples $\hat{x}$) leading to higher rewards.

Based on the Policy Gradient theorem, CA-GAN integrates reinforcement-driven
learning objective to guide the generator to preserve the causal structure.
To compute $\log p_{\theta}(\hat{x})$ in Eq (\ref{eq:causal-loss-derivative}),
CA-GAN treats the generator $G_{\theta}$ as a stochastic policy and
allows gradients to flow through \textit{log-probabilities} of its
outputs. Because real-world tabular datasets often contain both continuous
and categorical variables, CA-GAN adopts a mixed-type likelihood design
as in prior studies \cite{Xu2019,wen2022causal}.

For each categorical variable $X_{j}$ with $K$ possible categories,
the corresponding sub-generator $G_{j}$ outputs a vector of logits
$\boldsymbol{g}_{j}$. We then apply \textit{Gumbel-Softmax relaxation}
to these logits to enable differentiable sampling. With a temperature
$\tau>0$, we draw a stochastic, soft one-hot vector $\boldsymbol{a}_{j}$:
\begin{equation}
\boldsymbol{a}_{j}=\text{Gumbel-Softmax}(\boldsymbol{g}_{j},\tau),\label{eq:categorical-variable-stochastic}
\end{equation}
where $\boldsymbol{g}_{j}=G_{j}(\mathbf{PA}_{j},z_{j})$.

For each continuous variable $X_{j}$, we treat its corresponding
sub-generator output as the mean of Gaussian distribution with standard
deviation of 1, and draw a stochastic value $a_{j}$:
\begin{equation}
a_{j}\sim\mathcal{N}(G_{j}(\mathbf{PA}_{j},z_{j}),1)\label{eq:continuous-variable-stochastic}
\end{equation}
These sampling formulas introduce a controlled stochasticity into
the sub-generator outputs, which allows exploration of nearby values
in the output space. Each stochastic value $a_{j}$ and stochastic
vector $\boldsymbol{a}_{j}$ are associated with their log-probabilities
$\log p_{j}(a_{j})$ and $\log p_{j}(\boldsymbol{a}_{j})$. By summing
over all variables, we obtain the data-point-level log-likelihood:
\begin{equation}
\log p_{\theta}(\hat{x})=\sum_{j=1}^{M_{con}}\log p_{j}(a_{j})+\sum_{j=1}^{M_{cat}}\log p_{j}(\boldsymbol{a}_{j}),\label{eq:log-prob}
\end{equation}
where $M_{con}$ and $M_{cat}$ are the numbers of continuous and
categorical variables.

We combine the differentiable log-probability $\log p(\hat{x})$ in
Eq (\ref{eq:log-prob}) with the non-differentiable structural reward
$R(\hat{x})$ in Eq (\ref{eq:structural-reward}) to effectively optimize
the causal loss in Eq (\ref{eq:causal-loss-derivative}). To mitigate
the high variance in the policy gradient, we center the reward signal
by subtracting a moving-average baseline of past rewards. This stabilized
causal loss propagates reliable feedback to the generator, enabling
it to learn causal fidelity alongside statistical realism.

\section{Experiments}

\subsection{Experiment settings}

\begin{figure}[th]
\begin{centering}
\subfigure[4-node]{\includegraphics[scale=0.4]{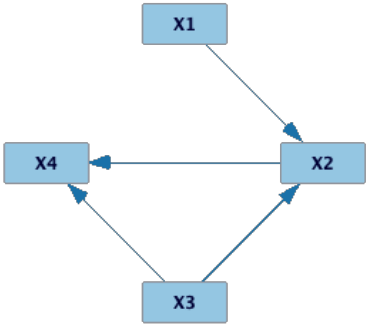}}\hspace{1cm}\subfigure[5-node]{\includegraphics[scale=0.4]{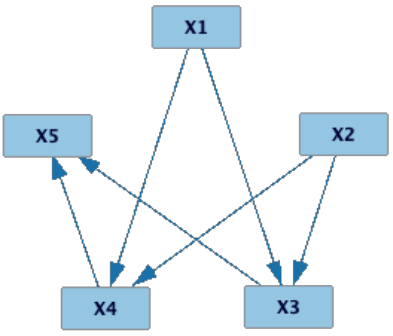}}\hspace{1cm}\subfigure[6-node]{\includegraphics[scale=0.4]{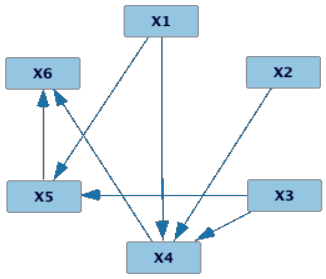}}
\par\end{centering}
\caption{\label{fig:Causal-graphs-(DAGs)}True causal graphs (DAGs) used to
construct 6 synthetic benchmark datasets.}
\end{figure}

\textbf{Datasets: }We evaluate CA-GAN on 14 tabular datasets, including
six synthetic datasets and eight real-world datasets. The synthetic
benchmark datasets are generated from known causal DAGs constructed
using TETRAD software \cite{scheines1998tetrad}, with the DAGs shown
in Figure \ref{fig:Causal-graphs-(DAGs)}. For each DAG, we produce
datasets with 10K and 20K samples. For real-world benchmark datasets
(Table \ref{tab:real-datasets}), we use public datasets that are
widely used for classification tasks across domains such as healthcare,
finance, and social science \cite{Xu2019,Borisov2023,zhang2023generative,nguyen2024tabular}.

\begin{table}[th]
\caption{\label{tab:real-datasets}Characteristics of 8 real-world benchmark
datasets.}

\centering{}\resizebox{0.6\textwidth}{!}{%
\begin{tabular}{|l|r|r|l|}
\hline 
\rowcolor{header_color}\textbf{Dataset} & \textbf{\# of samples $N$} & \textbf{\# of features $M$} & \textbf{Source}\tabularnewline
\hline 
\hline 
\textit{breast\_cancer} & 569 & 30 & scikit-learn\tabularnewline
\hline 
\rowcolor{even_color}\textit{diabetes} & 768 & 8 & OpenML\tabularnewline
\hline 
\textit{australian} & 690 & 14 & OpenML\tabularnewline
\hline 
\rowcolor{even_color}\textit{compas} & 4,010 & 10 & UCI\tabularnewline
\hline 
\textit{phoneme} & 5,404 & 5 & OpenML\tabularnewline
\hline 
\rowcolor{even_color}\textit{bank} & 4,521 & 14 & UCI\tabularnewline
\hline 
\textit{german} & 1,000 & 20 & UCI\tabularnewline
\hline 
\rowcolor{even_color}\textit{insurance} & 1,338 & 6 & Kaggle\tabularnewline
\hline 
\end{tabular}}
\end{table}

\textbf{Evaluation metrics: }We evaluate the generated data with the
following metrics:

\textit{Structural Hamming Distance }(SHD,\textit{ $\downarrow$}): This
metric is used in six synthetic benchmark datasets to assess \textit{causal
preservation}. A synthetic causal graph is extracted from the synthetic
samples using the PC algorithm. SHD score is computed as the number
of edge additions, deletions, or reversals needed to transform this
synthetic causal graph into the true causal graph (Figure \ref{fig:Causal-graphs-(DAGs)}).
We further report \textit{Global Conditional Independence} (GCI, $\uparrow$)
\cite{Jiang2026} and \textit{Average Treatment Effect} (ATE error,
$\downarrow$) as interventional metrics.

\textit{F1-score} (F1, $\uparrow$): This metric measures \textit{utility
preservation} i.e., how well ML classifiers trained on synthetic data
generalize to real data. For real-world benchmark datasets, we split
the data into 80\% $\mathcal{D}_{train}$ for training and 20\% for
testing $\mathcal{D}_{test}$. Tabular generation methods are trained
on ${\cal D}_{train}$ to generate a synthetic dataset ${\cal D}_{fake}$
such that $|{\cal D}_{fake}|=|{\cal D}_{train}|$. Then, an XGBoost
classifier is trained on $\mathcal{D}_{fake}$ and evaluated on the
real test set $\mathcal{D}_{test}$.

\textit{Distance to Closest Record }(DCR, $\uparrow$) \cite{yang2025doubling}:
This metric evaluates \textit{privacy preservation}. Given a fake
sample $\hat{x}\in{\cal D}_{fake}$, $\text{DCR}(\hat{x})=\min\{d(\hat{x},x_{i})\mid x_{i}\in\mathcal{D}_{train}\}$,
where $d(\cdot)$ is the Euclidean distance. As DCR measures how close
synthetic samples are to real ones, \textit{very low values indicate
privacy leakage while very high values suggest poor realism}.

\textit{Re-identification risk} (Risk, $\downarrow$) \cite{Yoon2020}:
The privacy of a synthetic dataset is measured by a simulated re-identification
attack. For each real sample $x_{i}\in\mathcal{D}_{real}$, we compute
its proximity to the synthetic dataset ${\cal D}_{fake}$ as $\hat{r}_{i}=\min_{\hat{x}_{j}\in{\cal D}_{fake}}||\boldsymbol{w}\cdot(x_{i}-\hat{x}_{j})||$,
where $\boldsymbol{w}$ is a weight vector for the weighted Euclidean
distance computation. We also compute its distinctness threshold in
the real dataset as $r_{i}=\min_{x_{j}\in\mathcal{D}_{real}\setminus x_{i}}||\boldsymbol{w}\cdot(x_{i}-x_{j})||$.
The re-identidication risk is defined as $\text{Risk}(\mathcal{D}_{real},{\cal D}_{fake})=\frac{1}{N}\sum_{i=1}^{N}\mathbb{I}(\hat{r}_{i}<r_{i})$,
where $\mathbb{I}$ denotes the indicator function. A lower score\textit{
}indicates stronger privacy protection.

\textit{Nearest Neighbor Distance Ratio }(NNDR, $\uparrow$) \cite{Ganev2025}:
This metric captures the similarity in local neighborhood structures
between real and synthetic datasets. For each synthetic sample $\hat{x}$,
the NNDR score is $\text{NNDR}(\hat{x})=\frac{d(\hat{x},NN_{1})}{d(\hat{x},NN_{2})},$
where $NN_{1}$ and $NN_{2}$ represent the first and second nearest
neighbors of $\hat{x}$ in the real dataset ${\cal D}_{train}$, respectively.
$d(\cdot)$ denotes the Euclidean distance. A higher score indicates
that the synthetic dataset better captures the intrinsic geometry
of the real dataset, reflecting high fidelity and diversity without
mode collapse.

\textbf{Baselines:} We compare CA-GAN with seven SOTA tabular generation
methods: \textit{CTGAN} \cite{Xu2019}, \textit{TVAE} \cite{Xu2019},
\textit{TabDiff} \cite{Shi2025}, \textit{Great} \cite{Borisov2023}, \textit{TapTap} \cite{zhang2023generative},
\textit{Pred-LLM} \cite{nguyen2024tabular}, and \textit{Causal-TGAN} \cite{wen2022causal}.
As only Causal-TGAN considers causal structure in its sampling phase,
it is our main competitor. For reproducibility, we use the baselines'
source codes.

For reference, the \textit{Original} method denotes the baseline where
the ML classifier is trained directly on the real dataset $\mathcal{D}_{train}$.
Following \cite{zhang2023generative,nguyen2024tabular}, we employ XGBoost \cite{chen2016xgboost}
for the ML classifier. More importantly, it is one of the most effective
and widely used algorithms for tabular classification \cite{shwartz2022tabular}.
We report the average score of each method with \textit{three random
runs}.

\textbf{Implement details: }Following \cite{Xu2019,wen2022causal},
we train CA-GAN with epochs $=300$, batch size $=500$, and noise
dimension $=16$. We also set $\lambda=0.01$ for Eq (\ref{eq:generator-loss}).
We run the PC algorithm with $\alpha=0.05$ to extract causal graphs.
Each sub-generator $G_{j}$ is a 4-layer multilayer perceptron (MLP)
that receives as input the concatenation of the parent variables $\mathbf{PA}_{j}$
and a latent noise vector $z_{j}\sim\mathcal{N}(0,1)$. The network
applies the LeakyReLU activations with a negative slope of 0.2 and Batch
Normalization with momentum of 0.8 after the second and third layers
to enhance stability during training. The hidden layers progressively
expand from 64 to 128 units. The output layer depends on the variable
type: a \textit{tanh activation} for a continuous variable and a \textit{Gumbel-Softmax layer}
for a categorical variable. The discriminator $D$ is a 3-layer MLP.
The first two hidden layers use LeakyReLU activations with a slope
of 0.2, each of them contains 256 units, followed by a linear output
layer. Both $G$ and $D$ are optimized using the Adam optimizer with
a learning rate of $2\times10^{-4}$, $\beta_{1}=0.5$, and $\beta_{2}=0.9$.

\subsection{Results and discussions}

\textbf{Causal preservation: }Table \ref{tab:shd-result} shows the
causal preservation performance. CA-GAN achieves the lowest average
SHD score (3.78), outperforming all competing methods. Causal-TGAN delivers
better performance than other generative models. However, CA-GAN further
improves over Causal-TGAN thanks to its integration of the inferred
causal graph into both training and sampling phases. Compared to the
Original method that discovers DAG from the real dataset, CA-GAN that
discovers DAG from the fake dataset achieves the same SHD scores for
half of the datasets. These results suggest that \textit{our generated
data highly preserve the causal structures in the original data}.

\begin{table}[th]
\caption{\label{tab:shd-result}Causal preservation performance of generation
methods. SHD $\pm$ (std) are reported on 6 synthetic datasets. \textbf{Bold}
and \uline{underline} denote the best and second-best.}

\centering{}\resizebox{\textwidth}{!}{{\renewcommand{\arraystretch}{1.2}%
\begin{tabular}{|l|ccccccccc|}
\hline 
\rowcolor{header_color}SHD ($\downarrow$) & Original & CTGAN & TVAE & TabDiff & Great & TapTap & Pred-LLM & Causal-TGAN & CA-GAN\tabularnewline
\hline 
\hline 
4nodes\_10k & 2.00 (0.0) & 3.67 (0.5) & 3.33 (0.5) & 4.67 (0.5) & 3.00 (0.0) & 3.00 (0.0) & 3.00 (0.0) & \textbf{2.00 (0.0)} & \textbf{2.00 (0.0)}\tabularnewline
\cline{1-1}
\rowcolor{even_color}5nodes\_10k & 0.00 (0.0) & 5.67 (0.5) & 5.67 (1.3) & 7.00 (0.8) & 7.33 (0.9) & \textbf{2.00 (0.0)} & 6.00 (0.0) & \uline{4.00 (0.0)} & \uline{4.00 (0.0)}\tabularnewline
6nodes\_10k & 5.00 (0.0) & 8.67 (1.3) & 8.33 (2.6) & 8.33 (1.9) & 9.67 (1.3) & 11.00 (0.0) & \textbf{5.00 (0.0)} & 6.00 (0.0) & \uline{5.67 (1.7)}\tabularnewline
\rowcolor{even_color}4nodes\_20k & 2.00 (0.0) & 3.00 (0.0) & 2.67 (0.9) & 4.00 (0.8) & 4.00 (0.0) & 5.00 (0.0) & 3.00 (0.0) & \textbf{2.00 (0.0)} & \textbf{2.00 (0.0)}\tabularnewline
5nodes\_20k & 4.00 (0.0) & 5.00 (0.8) & 4.67 (0.9) & 6.00 (0.0) & 5.67 (0.9) & 6.00 (0.0) & 6.00 (0.0) & \textbf{4.00 (0.0)} & \textbf{4.00 (0.0)}\tabularnewline
\rowcolor{even_color}6nodes\_20k & 4.00 (0.0) & 8.00 (0.8) & \uline{7.67 (1.7)} & 8.67 (2.1) & 8.33 (1.3) & 8.00 (0.0) & 9.00 (0.0) & 8.00 (0.8) & \textbf{5.00 (0.0)}\tabularnewline
\rowcolor{childheader_color}Average & 2.83 & 5.67 & 5.39 & 6.44 & 6.33 & 5.83 & 5.33 & \uline{4.33} & \textbf{3.78}\tabularnewline
\hline 
\end{tabular}}}
\end{table}

Beyond graph matching, we aggregate interventional metrics (GCI $\uparrow$
and ATE error $\downarrow$). CA-GAN achieves a higher average GCI
(0.62 vs. 0.60), indicating superior preservation of complex global
dependencies. Although Causal-TGAN yields a lower local ATE error
(0.83 vs. 1.48), this suggests that SHD and GCI primarily capture
global structural consistency, whereas ATE evaluates specific local
intervention effects. CA-GAN favors stronger overall causal structure
preservation while maintaining reasonable intervention fidelity.

\textbf{Utility preservation:} Table \ref{tab:F1-score-real} reports
F1-scores on eight real-world benchmark datasets. CA-GAN achieves
the best overall performance (0.6590), which closely matches the Original method
(0.6769). TapTap and Causal-TGAN perform comparatively well, ranking
just below CA-GAN. TapTap benefits from a tabular foundation model
while Causal-TGAN applies causal knowledge during sampling data. However,
CA-GAN consistently outperforms them. By incorporating causal information
into both training and sampling phases, our synthetic data better
respect causal constraints and deliver higher downstream task utility.

\begin{table}[!t]
\centering
\caption{\label{tab:F1-score-real}F1 $\pm$ (std) on 8 real-world datasets
to evaluate utility preservation performance.}

\centering{}\resizebox{\textwidth}{!}{{\renewcommand{\arraystretch}{1.2}%
\begin{tabular}{|l|ccccccccc|}
\hline 
\rowcolor{header_color}F1-score ($\uparrow$) & Original & CTGAN & TVAE & TabDiff & Great & TapTap & Pred-LLM & Causal-TGAN & CA-GAN\tabularnewline
\hline 
\hline 
breast\_ca & 0.972 (0.0) & 0.660 (0.2) & 0.942 (0.0) & 0.673 (0.0) & 0.710 (0.1) & \uline{0.958 (0.0)} & 0.954 (0.0) & 0.927 (0.1) & \textbf{0.961 (0.0)}\tabularnewline
\rowcolor{even_color}diabetes & 0.604 (0.0) & 0.375 (0.1) & \uline{0.633 (0.0)} & 0.181 (0.2) & 0.360 (0.1) & 0.621 (0.0) & 0.562 (0.0) & 0.618 (0.1) & \textbf{0.640 (0.1)}\tabularnewline
australian & 0.858 (0.0) & 0.353 (0.2) & 0.757 (0.0) & 0.424 (0.1) & 0.693 (0.0) & \textbf{0.859 (0.0)} & 0.834 (0.0) & 0.823 (0.1) & \uline{0.845 (0.0)}\tabularnewline
\rowcolor{even_color}compas & 0.278 (0.0) & 0.116 (0.0) & 0.210 (0.1) & 0.028 (0.0) & 0.094 (0.0) & \textbf{0.323 (0.0)} & 0.257 (0.0) & 0.250 (0.0) & \uline{0.260 (0.0)}\tabularnewline
phoneme & 0.824 (0.0) & 0.444 (0.0) & 0.642 (0.0) & 0.207 (0.1) & 0.556 (0.1) & 0.687 (0.0) & 0.751 (0.0) & \uline{0.765 (0.0)} & \textbf{0.772 (0.0)}\tabularnewline
\rowcolor{even_color}bank & 0.400 (0.0) & 0.075 (0.0) & 0.319 (0.1) & 0.006 (0.0) & 0.061 (0.0) & 0.289 (0.1) & 0.307 (0.1) & \uline{0.347 (0.0)} & \textbf{0.357 (0.0)}\tabularnewline
german & 0.554 (0.1) & 0.371 (0.0) & 0.455 (0.1) & 0.164 (0.2) & 0.294 (0.0) & 0.428 (0.1) & 0.220 (0.1) & \uline{0.457 (0.0)} & \textbf{0.498 (0.0)}\tabularnewline
\rowcolor{even_color}insurance & 0.925 (0.0) & 0.525 (0.2) & 0.819 (0.1) & 0.627 (0.0) & 0.697 (0.1) & 0.920 (0.0) & 0.922 (0.0) & \uline{0.929 (0.0)} & \textbf{0.939 (0.0)}\tabularnewline
\hline 
\rowcolor{childheader_color}Average & 0.6769 & 0.3649 & 0.5971 & 0.2887 & 0.4330 & 0.6355 & 0.6006 & \uline{0.6395} & \textbf{0.6590}\tabularnewline
\hline 
\end{tabular}}}
\end{table}

\begin{figure}[th]
\begin{centering}
\includegraphics[width=0.8\columnwidth]{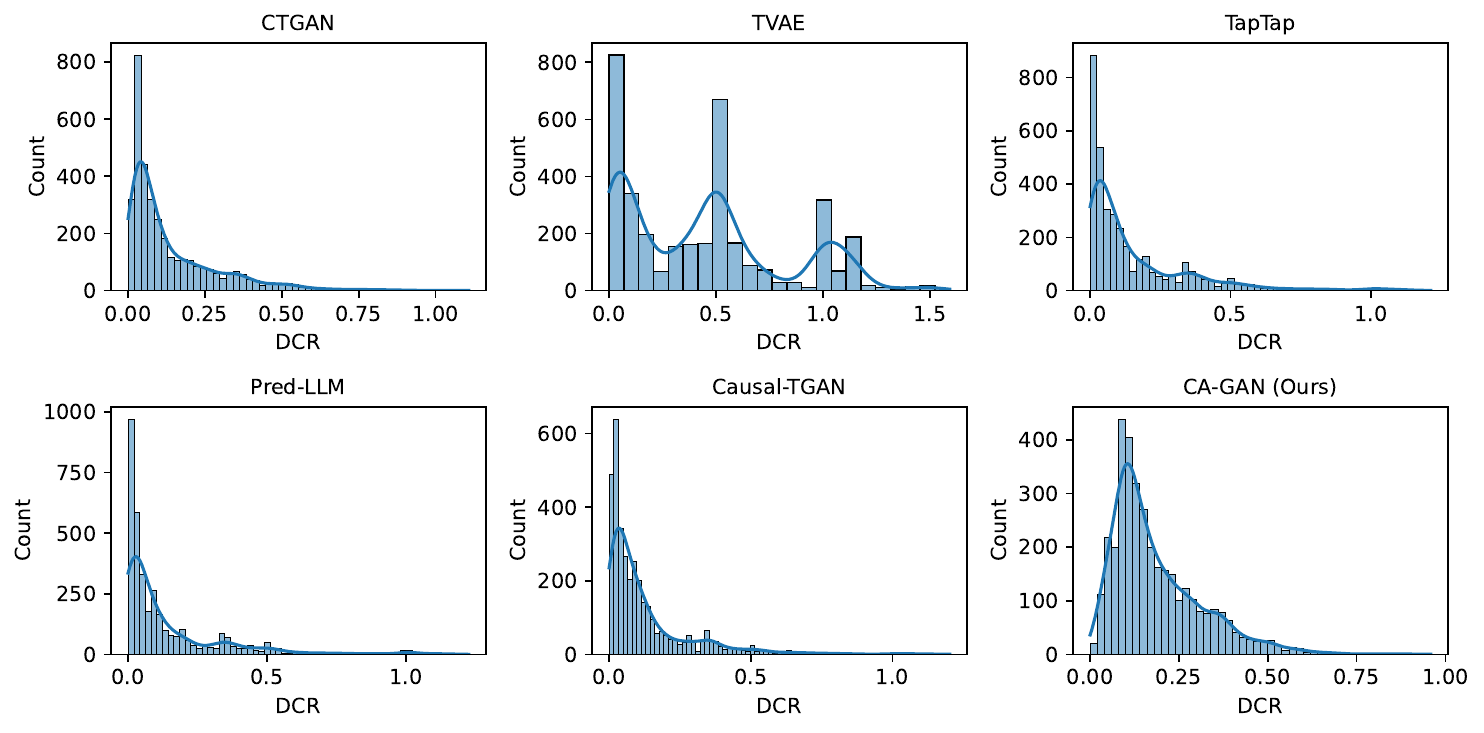}
\par\end{centering}
\caption{\label{tab:DCR}DCR distributions on the dataset \textit{bank}. A
very small DCR suggests the method may copy certain feature values
from the original data (i.e., potential information leakage), whereas
a very high DCR indicates unrealistic generated samples.}
\end{figure}

\textbf{Privacy preservation:} Figure \ref{tab:DCR} shows the DCR
distribution of each method on the \textit{bank} dataset. Other methods
e.g., CTGAN, TapTap, Pred-LLM, and Causal-TGAN have sharp peaks (i.e.,
modes) near $[0,0.05]$, meaning their synthetic samples are too close
to the real samples, which may cause privacy leakage. In contrast, CA-GAN has
a smoother curve with its main peak around $[0.1,0.15]$, showing
that our generated samples are similar to the real samples but not
identical. This helps prevent direct copying of real records and protects
privacy. Although TVAE produces higher DCR values, many of them might
be outliers, leading to unrealistic synthetic samples and lower usefulness.
As shown in Table \ref{tab:F1-score-real}, the F1-score of TVAE on
this dataset is 0.319 (4\% lower than ours).

\begin{figure}[th]
\centering{}\subfigure[Risk score ($\downarrow$)]{\includegraphics[width=0.4\columnwidth]{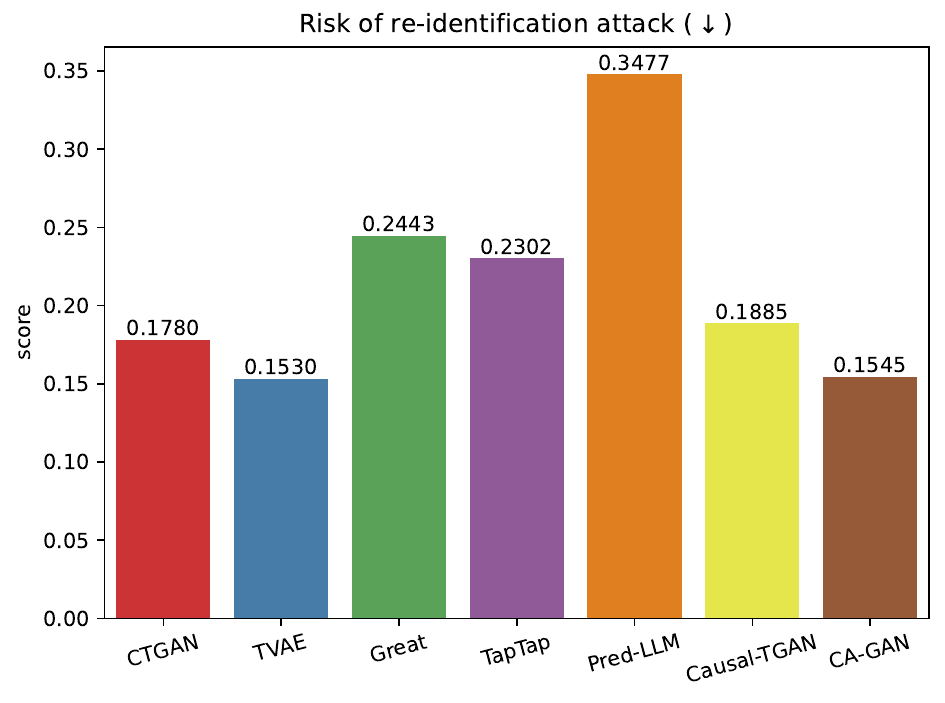}}\hspace{1cm}\subfigure[Quality score ($\uparrow$)]{\includegraphics[width=0.4\columnwidth]{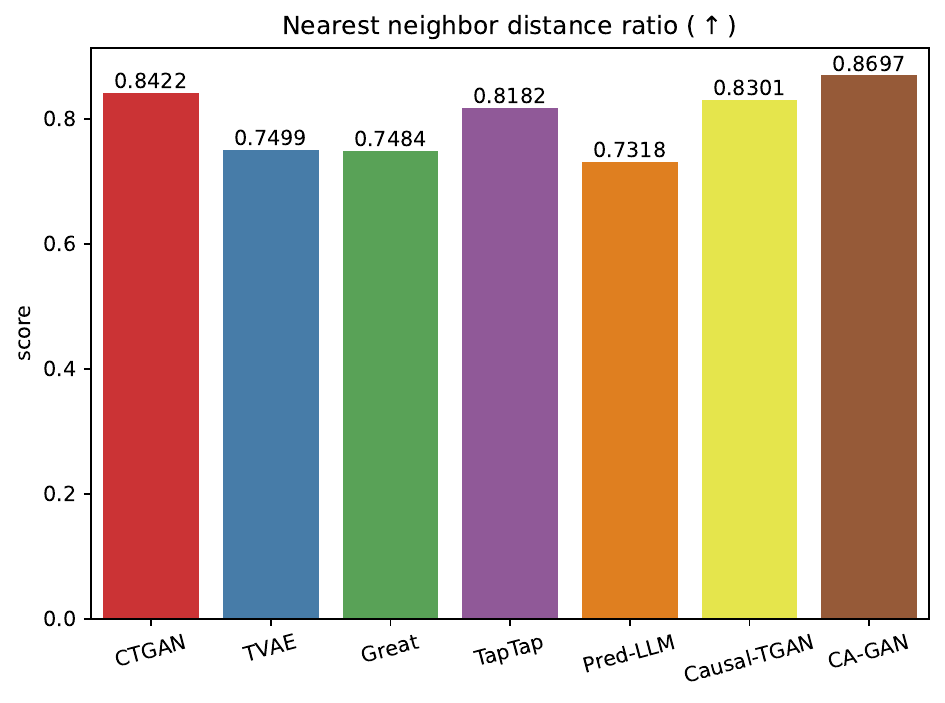}}\caption{\label{fig:Comparison-of-privacy-quality}Average re-identification
risk and quality score over 8 real-world datasets.}
\end{figure}

\textbf{Re-identification risk:} The defense results in Figure \ref{fig:Comparison-of-privacy-quality}(a)
further validates the privacy robustness of CA-GAN. We observe that
GAN-based methods achieve lower re-identification risks than LLM-based
methods, reflecting stronger resistance to privacy leakage. Among
them, CA-GAN attains the lowest risk of 0.1545, which highlights its
ability to effectively mitigate the risk of re-identification attacks
and improves privacy preservation.

\textbf{Data quality:} Figure \ref{fig:Comparison-of-privacy-quality}(b)
presents the NNDR results, which measure how well synthetic samples
capture the local structure of real data. CA-GAN achieves the highest
NNDR score (0.8697), outperforming all baselines. CTGAN and Causal-TGAN also
show relatively strong performance while the scores of TVAE, Great,
and Pred-LLM are much lower, suggesting their weaker local consistency.

Overall, CA-GAN keeps a good balance between privacy and realism,
creating synthetic data that stay close enough to real patterns without
risking information leakage.

\subsection{Ablation studies}

\begin{figure}[th]
\begin{centering}
\includegraphics[width=0.7\columnwidth]{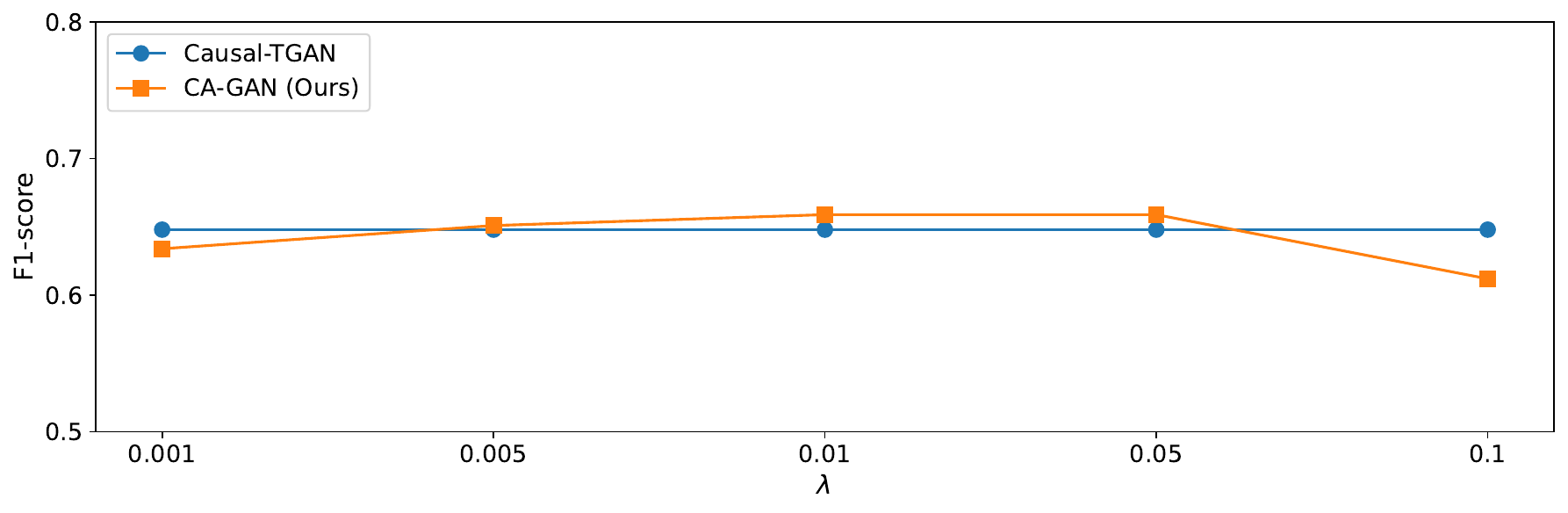}
\par\end{centering}
\caption{\label{fig:coefficient-lambda}Influence of $\ensuremath{\lambda}$
on CA-GAN's performance over 8 real-world datasets.}
\end{figure}

\textbf{Loss coefficient $\lambda$:} We analyze the impact of the
coefficient $\ensuremath{\lambda}$ value in Eq (\ref{eq:generator-loss}),
which balances the causal loss and the adversarial loss. Since the
SHD-based reward can dominate training when its magnitude grows with
the number of features $(M\times M)$, $\ensuremath{\lambda}$ is
introduced to control this influence and maintain stability. We set
a small value for $\lambda$ as the causal loss is much higher than
the adversarial loss. From Figure \ref{fig:coefficient-lambda}, CA-GAN
achieves the best performance when $\ensuremath{\lambda}$ is in $[0.01,0.05]$
i.e., the causal and adversarial losses are well balanced. Setting
$\ensuremath{\lambda}$ too high makes the causal loss overpower the
adversarial loss whereas setting it too low allows the adversarial
loss to dominate. In both cases, our performance drops. Compared to
Causal-TGAN, our method is always better with a small enough value
for $\lambda$.

\begin{figure}[th]
\begin{centering}
\includegraphics[width=0.8\columnwidth]{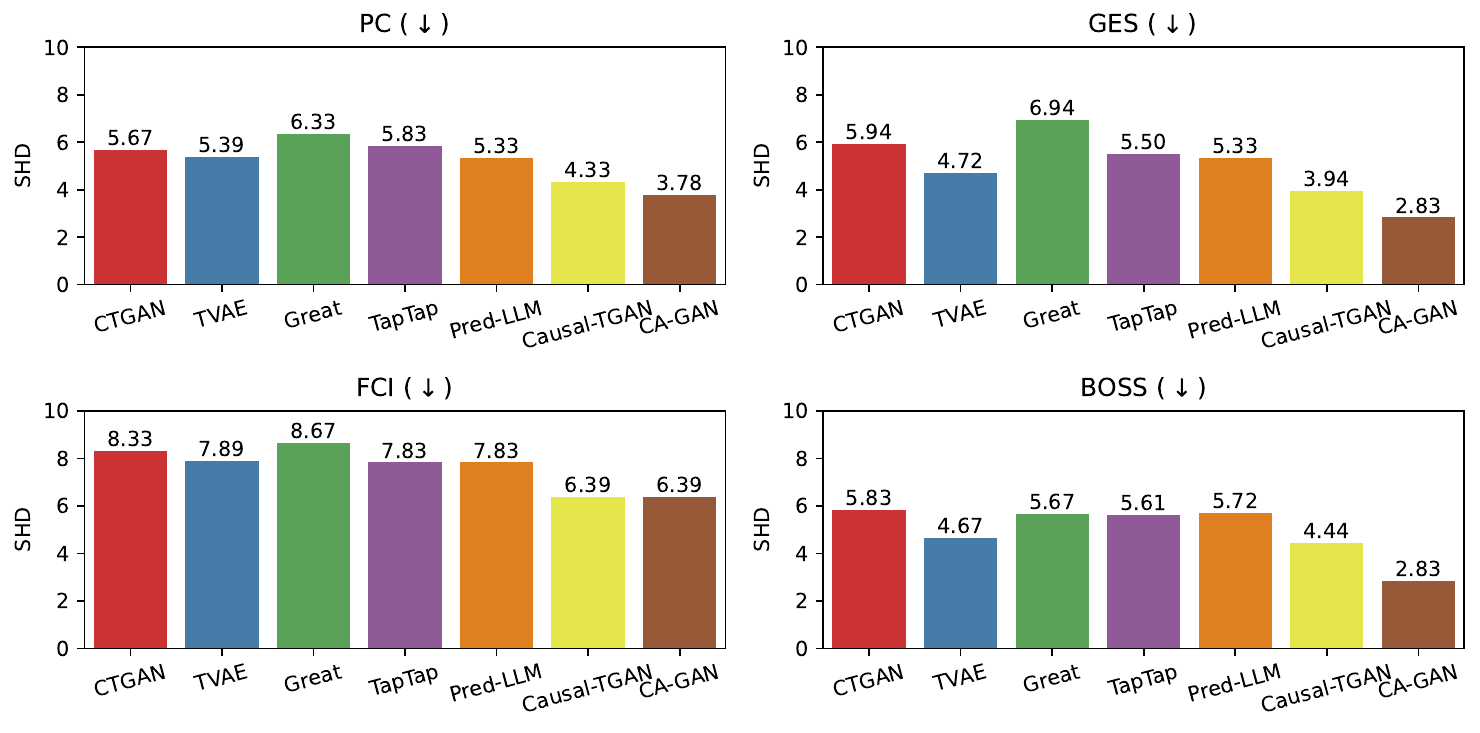}
\par\end{centering}
\caption{\label{fig:Ablation-study-CD}The impact of different causal discovery
methods. We report the average SHD ($\downarrow$) over 6 synthetic
datasets. The consistent ranking across PC, GES, FCI, and BOSS demonstrates
that CA-GAN's benefit is not tied to any particular CD method.}
\end{figure}

\textbf{Different causal discovery (CD) methods:} In our causal preservation
experiment, we use PC to extract synthetic causal graph and compare
them with the ground-truths. To check whether CA-GAN\textquoteright s
causal fidelity is sensitive to a particular CD method, we repeat
the same evaluation with three additional CD methods, namely \textit{Greedy
Equivalence Search} (GES) \cite{Chickering2002} -- a score-based
method, \textit{Fast Causal Inference} (FCI) \cite{Spirtes2013} --
a constraint-based method, and \textit{Best Order Score Search} (BOSS)
\cite{Andrews2023} -- a permutation-based method. From Figure \ref{fig:Ablation-study-CD},
CA-GAN consistently achieves the lowest SHD scores across all CD methods,
demonstrating its strong adaptability and causal fidelity. As expected,
Causal-TGAN is the second-best method as it also considers the causal
relationship during sampling data. However, CA-GAN is consistently
better than Causal-TGAN. These results confirm that our RL--based
training objective effectively preserves causal relationships in the
original data regardless of the CD algorithm employed.

\textbf{Robustness to causal graph:}\textit{ }We evaluate whether
CA-GAN is overly dependent on the estimated causal graph $\mathcal{G}_{real}$.
We simulate a 10K-sample dataset with the \textit{ground-truth} DAG:
\textbf{$A\rightarrow S\leftarrow E$}, then systematically inject
structural errors. Using these corrupted DAGs as the prior DAG $\mathcal{G}_{real}$,
we evaluate the SHD between the DAG recovered from generated data
and the ground-truth DAG. Table \ref{tab:quality-graph} shows that
CA-GAN remains stable under moderate graph error. When the input graph
becomes increasingly corrupted, CA-GAN degrades more gracefully than
Causal-TGAN e.g., CA-GAN obtains lower SHD under 40-60\% errors. These
results suggest that CA-GAN does not treat $\mathcal{G}_{real}$ as
a hard constraint; instead, ${\cal G}_{real}$ acts as a soft structural
prior. Moreover, it indicates CA-GAN's resilience to the imperfect
graphs typically generated by causal discovery methods in practice.

\begin{table}[t]
\centering
\caption{\label{tab:quality-graph}DAG-error and SHD ($\downarrow$) of Causal-TGAN
and CA-GAN.}

\centering{}\resizebox{0.6\textwidth}{!}{%
\begin{tabular}{|l|l|r|r|}
\hline 
\rowcolor{header_color}\textbf{DAG errors (\%)} & \textbf{$\mathcal{G}_{real}$} & \textbf{Causal-TGAN} & \textbf{CA-GAN}\tabularnewline
\hline 
0 (True mechanism) & $A\rightarrow S\leftarrow E$ & 2.00 & 2.00\tabularnewline
\hline 
\rowcolor{even_color}20 (Missed parent) & $A\rightarrow S\hspace{0.6cm}E$ & 2.00 & 2.00\tabularnewline
\hline 
40 (No edges found) & $A\hspace{0.6cm}S\hspace{0.6cm}E$ & 2.33 & 2.00\tabularnewline
\hline 
\rowcolor{even_color}60 (Wrong chain) & $A\rightarrow E\rightarrow S$ & 2.67 & 2.33\tabularnewline
\hline 
80 (Effect as cause) & $A\leftarrow S\hspace{0.6cm}E$ & 3.00 & 3.00\tabularnewline
\hline 
\rowcolor{even_color}100 (Total mismatch) & $A\leftarrow S\rightarrow E$ & 3.00 & 3.00\tabularnewline
\hline 
\end{tabular}}
\end{table}

\begin{figure}[h]
\centering
\includegraphics[width=0.8\textwidth]{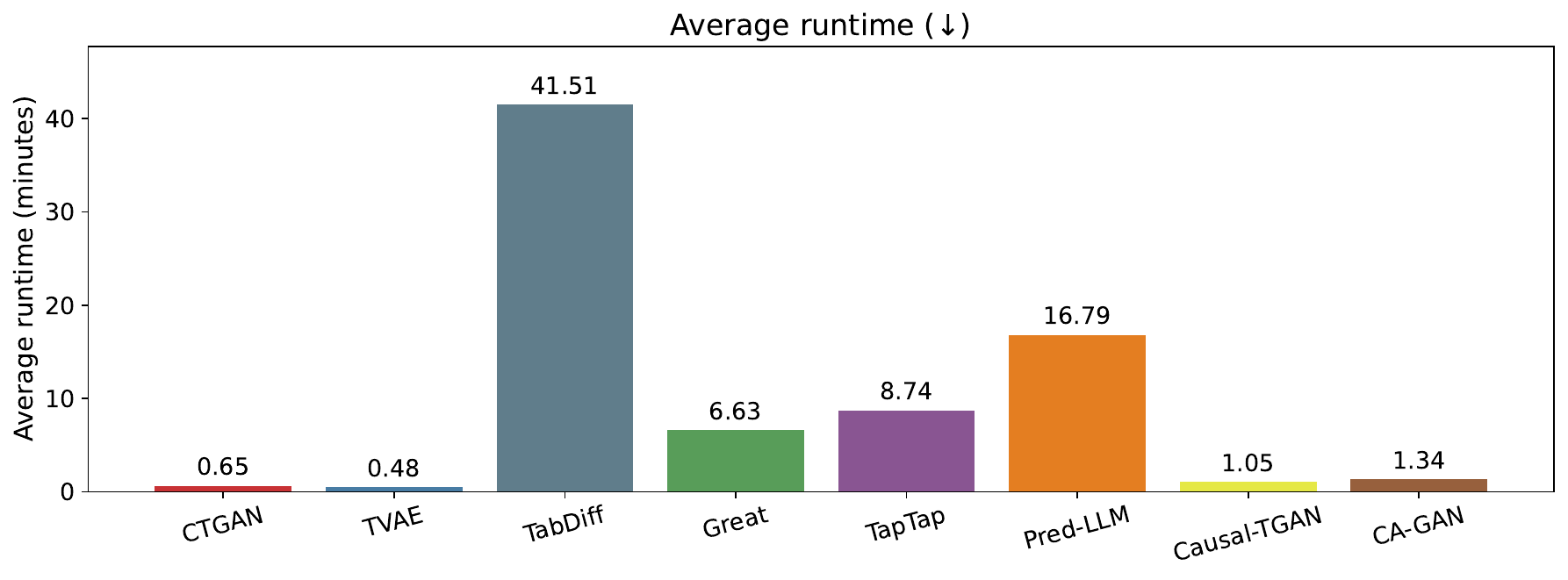}

\caption{\label{fig:Average-runtime}Average runtime (\textbf{in minutes})
of baselines across eight real-world datasets.}
\end{figure}

\textbf{Computational efficiency: }Figure \ref{fig:Average-runtime}
compares the average runtime across eight real-world datasets on an\textit{
NVIDIA GeForce RTX 4070 Super}. CA-GAN achieves competitive computational
efficiency, averaging 1.34 minutes per dataset, while improving causal
fidelity, downstream performance, privacy, and data quality.

\section{Conclusion}

We address the challenge of generating synthetic tabular data that
retain both \textit{statistical realism} and \textit{causal integrity}.
These two aspects are rarely achieved simultaneously by existing methods.
Different from other methods, our method focuses on causality-aware
training through reinforcement learning, supported by causal graph
extraction and graph-conditioned generative architecture to ensure
faithful causal preservation in synthetic tabular data. We present
CA-GAN, a causal-aware generative framework that integrates the causal
knowledge into the synthesis process to produce data consistent with
the underlying causal mechanisms. We conduct comprehensive experiments
on 14 datasets and compare against seven recent SOTA baselines. We
demonstrate that CA-GAN achieves the best performance across causal
preservation, downstream task, data quality, and privacy protection.

\bibliography{references}

\end{document}